\documentclass[sigconf]{acmart}

\AtBeginDocument{%
  \providecommand\BibTeX{{%
    \normalfont B\kern-0.5em{\scshape i\kern-0.25em b}\kern-0.8em\TeX}}}

\copyrightyear{2022}
\acmYear{2022}
\setcopyright{acmcopyright}\acmConference[SIGIR '22]{Proceedings of the 45th
International ACM SIGIR Conference on Research and Development in Information
Retrieval}{July 11--15, 2022}{Madrid, Spain}
\acmBooktitle{Proceedings of the 45th International ACM SIGIR Conference on
Research and Development in Information Retrieval (SIGIR '22), July 11--15, 2022,
Madrid, Spain}
\acmPrice{15.00}
\acmDOI{10.1145/3477495.3531746}
\acmISBN{978-1-4503-8732-3/22/07}

\usepackage[T1]{fontenc}
\usepackage[utf8]{inputenc}
\usepackage{amsmath}
\usepackage{subfigure}
\usepackage{longtable}
\usepackage{booktabs} 
\usepackage{multirow}
\usepackage{multicol}
\usepackage{color}

\usepackage{balance}

\newcommand\tf[1]{\textbf{#1}}
\newcommand{\ours}{KnowPrompt}
\newcommand{\ourss}{RetrievalRE}

\newcommand{\knn}{$k$NN}

\begin{document}
\fancyhead{} 

\title{Relation Extraction as Open-book Examination:\\ Retrieval-enhanced Prompt Tuning}




\author{Xiang Chen}
\authornote{Equal contribution.}
\affiliation{%
  \institution{Zhejiang University \\ AZFT Joint Lab for Knowledge Engine \\ Hangzhou Innovation Center}
  \city{Hangzhou}
  \state{Zhejiang}
  \country{China}
}
\email{xiang_chen@zju.edu.cn}

\author{Lei Li}
\authornotemark[1]
\affiliation{%
  \institution{Zhejiang University \\ AZFT Joint Lab for Knowledge Engine \\ Hangzhou Innovation Center}
  \city{Hangzhou}
  \state{Zhejiang}
  \country{China}
}
\email{leili21@zju.edu.cn}

\author{Ningyu Zhang}
\authornotemark[2]
\affiliation{%
  \institution{Zhejiang University \\ AZFT Joint Lab for Knowledge Engine \\ Hangzhou Innovation Center}
  \city{Hangzhou}
  \state{Zhejiang}
  \country{China}
}
\email{zhangningyu@zju.edu.cn}

\author{Chuanqi Tan}
\affiliation{%
  \institution{Alibaba Group}
  \city{Hangzhou}
  \state{Zhejiang}
  \country{China}
}
\email{chuanqi.tcq@alibaba-inc.com}

\author{Fei Huang}
\author{Luo Si}
\affiliation{%
  \institution{Alibaba Group}
  \city{Hangzhou}
  \state{Zhejiang}
  \country{China}
}
\email{f.huang@alibaba-inc.com}
\email{luo.si@alibaba-inc.com}


\author{Huajun Chen}
\authornote{Corresponding author.}
\affiliation{%
  \institution{Zhejiang University \\ AZFT Joint Lab for Knowledge Engine \\ Hangzhou Innovation Center}
  \city{Hangzhou}
  \state{Zhejiang}
  \country{China}
}
\email{huajunsir@zju.edu.cn}


\begin{abstract}

Pre-trained language models have contributed significantly to relation extraction by demonstrating remarkable few-shot learning abilities. However, prompt tuning methods for relation extraction may still fail to generalize to those rare or hard patterns. Note that the previous parametric learning paradigm can be viewed as {\it memorization} regarding training data as a book and inference as the close-book test. Those long-tailed or hard patterns can hardly be memorized in parameters given few-shot instances. To this end, we regard RE as an open-book examination and propose a new semiparametric paradigm of retrieval-enhanced prompt tuning for relation extraction. We construct an open-book datastore for retrieval regarding prompt-based instance representations and corresponding relation labels as memorized key-value pairs. During inference, the model can infer relations by linearly interpolating the base output of PLM  with the non-parametric nearest neighbor distribution over the datastore. In this way, our model not only infers relation through knowledge stored in the weights during training but also assists decision-making by unwinding and querying examples in the open-book datastore. Extensive experiments on benchmark datasets show that our method can achieve state-of-the-art in both standard supervised and few-shot settings\footnote{Code are available in \url{https://github.com/zjunlp/PromptKG/tree/main/research/RetrievalRE}.}.


\end{abstract}

\ccsdesc[500]{Computing methodologies~Information extraction}

 \keywords{Relation Extraction, Prompt Tuning, Few-shot Learning}


\maketitle

\section{Introduction}

Relation Extraction (RE) aims to detect the relations between the entities contained in a sentence, which has become a fundamental task for knowledge graph construction, benefiting many web applications, e.g., information retrieval~\cite{DBLP:conf/sigir/DietzKM18,DBLP:conf/sigir/Yang20}, recommender systems \cite{AliCG} and question answering~\cite{DBLP:conf/cikm/JiaPRW21,DBLP:conf/sigir/QuZ0CL21}.
With the rise of a series of pre-trained language models (PLMs)~\cite{devlin2018bert,DBLP:journals/corr/abs-1907-11692,DBLP:conf/nips/BrownMRSKDNSSAA20}, fine-tuning PLMs has become a dominating approach to RE~\cite{DBLP:journals/tacl/JoshiCLWZL20,ReadsRE,DBLP:journals/corr/abs-2102-01373,docunet}.
However, there exists a signiﬁcant objective gap between pre-training and ﬁne-tuning, which leads to performance decay in the low-data regime.

To address this issue, a new fine-tuning methodology named prompt tuning has been proposed, which makes waves in the community by demonstrating astounding few-shot capabilities on widespread tasks~\cite{DBLP:journals/corr/abs-2012-15723,schick2020automatically,liu2021gpt,knowledgeprompt,DBLP:journals/corr/abs-2201-08670,DBLP:journals/corr/abs-2201-07126}.
Concretely, with prompt tuning, the input texts are wrapped with task-specific templates to a prompt to reformulate input examples into cloze-style phrases (e.g. ``<\textit{input example}> It is $\texttt{[MASK]}.$ ''), and a verbalizer to map labels to candidate words (e.g.,  \textit{positive} $\rightarrow$ ``\textit{great}'' and \textit{negative} $\rightarrow$ ``\textit{terrible}'').
However, when applying prompt tuning to RE, vanilla prompt tuning methods will struggle with handling complex label verbalizations.
To this end, \citet{ptr} propose PTR for relation extraction, which applies logic rules to construct prompts with several sub-prompts. 
Meanwhile, \citet{chenKnowprompt2022} present KnowPrompt with learnable virtual answer words to represent rich semantic information of relation labels.

Fundamentally, previous RE methods with prompt tuning follow the parametric-based learning paradigm, which tries to maximize the probability of the golden relation with cross-entropy loss over their designed verbalizer during training, and then predict with the highest likelihood during inference. 
The regular training-test procedure can be regard as {\it memorization} if we view the training data as a {\it book} and inference as the {\it close-book examination}. 
Despite the success of prompt tuning PLMs for RE tasks, the existing {\it memorization-based} prompt tuning paradigm still suffers from the following limitations: 
the PLMs usually cannot generalize well for hard examples and perform unstably in an extremely low-resource setting since the scarce data or complex examples are not easy to be memorized in model embeddings during training.

To mitigate these issues, we take the first step to shift the closed-book test to open-book examination for RE; thus, the model can be provided with related instances as reference for decision-making.
Specifically, we propose retrieval-enhanced prompt tuning (\ourss), a new paradigm for RE, which empowers the model to refer to  similar instances from the training data and regard them as the cues for inference, to improve the robustness and generality when encountering extremely long-tailed or hard examples.
We construct the open-book datastore for RE by utilizing the embeddings of the ``$\texttt{[MASK]}$'' token in prompt for each instance. 
Then, during inference, the model retrieves Top-$k$ nearest reference instances as cues from the open-book datastore and makes inference by linearly interpolating the output of prompt-tuning with a non-parametric nearest neighbor distribution. 

To summarize, we build the first semi-parametric paradigm of retrieval-enhanced prompt tuning for RE, which can infer relation via explicit memorization.
Our work may open up new avenues for improving relation extraction through explicit memory.
Furthermore, we evaluate {\ourss} on popular RE benchmarks and demonstrate its advantageous over  baselines.
Finally, through detailed ablation and case study, we demonstrate the essential role of retrieval augmentation in boosting the RE performance, which might be easier to access via explicit memory.

\section{Methodology} 

\subsection{Prompt Tuning for RE}

Prompt-tuning is proposed to bridge the gap between pre-training and downstream tasks. 
Specifically, we denote $\mathcal{M}$, $\mathcal{T}$ and $\mathcal{\hat{V}}$ as the PLM, template function and verberlizer function for prompt tuning, respectively. Moreover, $s$ and $o$ represent the subject and object entity, respectively.
Formally, the RE task takes a query sentence $x$ with corresponding entity pair $(s,o)$ as input,
aiming to learn a distribution $P(r | (s, o))$  over all possible predefined relations $r \in \mathcal{R}$.

As for  RE with prompt tuning, given each instance $x$ for RE, the template function $\mathcal{T}$ inserts pieces of texts with entity pair $(s,o)$ into $x$ to map $x$  as $x_{prompt} = \mathcal{T}(x)$, where the $x_{prompt}$ is  the corresponding input of $\mathcal{M}$ with a {\tt[MASK]} token in it.
$V$ refers to the label words set, and $\mathcal{R}$ donates the relation set.
The verbalizer $\mathcal{\hat{V}}:\mathcal{R} \rightarrow V$ represents a mapping from relation label space to label word space, $\mathcal{\hat{V}}(r)$ means label words corresponding to label $r$.
Then  $\mathcal{M}$ produces the hidden vector at the \texttt{[MASK]} position $h_{\texttt{[MASK]}}$ to obtain instance representation $h_{x}$. 
The probability distribution over the relation set is calculated as:
    \begin{equation}
        P(r|x) =  P_\mathcal{M} ({\text{\tt [MASK]}} = v|\mathcal{T}(x))
    \end{equation},
where $v \in {V}$.
In this way, the RE problem can be transferred into a masked language modeling problem by filling the {\tt[MASK]} token in the input. 
To alleviate the issue of complicated verbalizer engineering, we follow \cite{warp,chenKnowprompt2022} to adopt a soft verbalizer for connecting the relation label to the label word by one-to-one mapping.

\subsection{Open-book Datastore for Retrieval}

As for the retrieval-enhanced prompt tuning, one core issue is constructing an external datastore for retrieval.
In this section, we introduce open-book datastore for retrieval, which contains fixed embeddings based on tuned PLMs rather than optimizing with the PLMs together. 
Typically, the construction of the open-book datastore mainly includes prompt-based instance representation and datastore collection.

\begin{figure}[!htbp] 
\centering 
\includegraphics[width=0.4\textwidth]{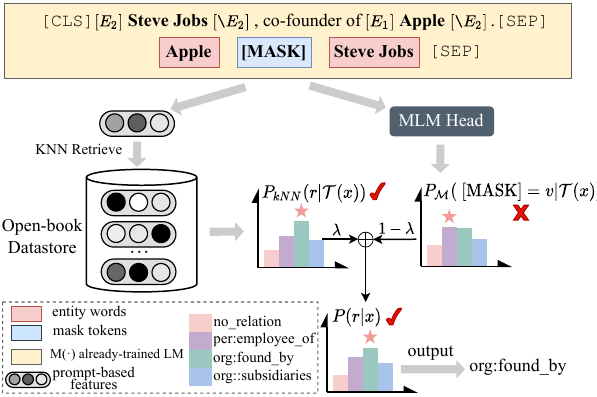} 
\caption{The inference procedure of our \ourss.} 
\label{fig:model}
\end{figure}

\subsubsection{{Prompt-based Instance Representation}}
We originally present prompt-based instance representation to obtain sentence embeddings. 
By leveraging the output embeddings of the ``$\texttt{[MASK]}$ '' token of the last layer of the underlying PLM, we can effectively exploit the explicit knowledge learned during pre-training.
For instance in Figure~\ref{fig:model}, we have a template ``[x] [s]  $\texttt{[MASK]}$  [o]'', where [x], [s] and [o] are the  placeholder for the  input sentences, subject entity and object entity respectively.
Given a instance $c$, we map $c$ to  $c_{prompt}$ with the template $\mathcal{T}(\cdot)$ 
and feed $c_{prompt}$ to the PLM $\mathcal{M}$ to generate the sequence representation $\mathcal{M}(\mathcal{T}(c))$. Then, we adopt the hidden vector of $\texttt{[MASK]}$ token as the prompt-based instance  representation $\mathbf{h_c}$ as:
\begin{equation}\label{eq:rep}
\mathbf{h}_{c} = \mathcal{M}(\mathcal{T}(c))_{\texttt{[MASK]}} .
\end{equation}

It is worth noting that our prompt-based instance representation can encode the semantic information of the instance and relation labels, consistent with the prompt learning.

\subsubsection{{Datastore Collection}}

We construct the datastore with the text collection $C$ from the training set.
Specifically, we utilize the prompt-based instance representation over the training set to construct the datastore.
Given the $i$-th example $\left(c_i,r_i\right)$ in the text collection $\mathcal{C}$,
we compute the key-value pair $(h_{c_i},r_i)$, in which  $h_{c_i}$ is a context embedding computed in Eq~\eqref{eq:rep}, and $r_i$ is the relation, respectively.
We construct the key-value datastore $\left(\mathcal{K},\mathcal{V}\right)$ with 
storing all pairs $(h_{c},r)$ and denote 
$h_{c}$ as \emph{key} and $r$ as \emph{value} as follows:
\begin{align}
\left(\mathcal{K},\mathcal{V}\right) = 
\{
\left(h_{c_i},r_i\right) \mid \left({c_i},r_i\right)\in \mathcal{C} 
\}
\end{align}
Note that the datastore is flexible to add, edit or delete any instances, thus, making it possible to post hoc model editing without re-training, and we leave this for future work.

\subsection{Retrieval-enhanced Relation Prediction}
The retrieval component is attached to the inference stage.
Given a query context $x$, we first  convert $x$ to $x_{prompt}$ and infer the model to obtain the $h_{x}$. 
Then  the model leverage the $h_{x}$ querying the open-book datastore $\left(\mathcal{K},\mathcal{V}\right)$ to retrieve the $k$-nearest neighbors $\mathcal{N}$ of $h_{x}$, according to a distance function $d(\cdot, \cdot)$, where
$d(\cdot, \cdot)$ typically adopt the Euclidean distance.
Then, we apply the softmax function over the negative distances to
compute the distribution over neighbors and aggregate probability mass to form the probability distribution over pre-defined relations:

\begin{align}
P_{\text{\knn{}}}\left(r \mid \mathcal{T}(x)\right)  \propto
 \sum_{\left(c_i,r_i\right)\in \mathcal{N}} {1}_{r=r_i} \exp\left(-d\left( h_x, h_{c_i}\right)\right) .
\label{eq:knnscore}
\end{align}
Afterwards, we reformulate the $P(r\mid x)$ by interpolating the non-parametric $k$ nearest neighbor distribution $P_{k\text{NN}}$ with the already-trained base PLM's MLM prediction $P_\mathcal{M}$ using parameter $\lambda$ to produce the final probability of the relation:

\begin{equation}
P(r \mid x)=\lambda P_{k\mathrm{NN}}(r \mid \mathcal{T}(x))+(1-\lambda)  P_\mathcal{M} ({\text{\tt [MASK]}} = v|\mathcal{T}(x)) .
\end{equation}

\begin{table*}[!t]
\centering
\small

\caption{
\label{tab:low-resource}
Model performance of RE models in the low-resource setting. We report the mean and standard deviation performance of micro $F_1$ scores (\%) over 5 different splits. The best numbers are highlighted in each column.}

\scalebox{0.85}{
\begin{tabular}{c|l|ccc|ccc|ccc}
\toprule

{\multirow{2}{*}{Source}} 
& {\multirow{2}{*}{Model}} 
& \multicolumn{3}{c|}{SemEval}
& \multicolumn{3}{c|}{TACRED}
& \multicolumn{3}{c}{TACREV}\\
\cmidrule{3-11}
 &  &K=1 &K=5 &K=16  &K=1 &K=5  &K=16  &K=1 &K=5 &K=16  \\

\midrule

    \multirow{5}{*}{None} 
   & \textsc{Fine-tuning}~\  
   & 18.5 \footnotesize{($\pm$ 1.4)} & 41.5 \footnotesize{($\pm$ 2.3)} & 66.1 \footnotesize{($\pm$ 0.4)}
   & 7.6 \footnotesize{($\pm$ 3.0)} & 16.6 \footnotesize{($\pm$ 2.1)} & 26.8 \footnotesize{($\pm$ 1.8)} 
   & 7.2 \footnotesize{($\pm$ 1.4)} & 16.3 \footnotesize{($\pm$ 2.1)} & 25.8 \footnotesize{($\pm$ 1.2)} \\
   
    & \textsc{GDPNet}~\  
   & 10.3 \footnotesize{($\pm$ 2.5)} 
   & 42.7 \footnotesize{($\pm$ 2.0)} 
   & 67.5 \footnotesize{($\pm$ 0.8)}
   & 4.2 \footnotesize{($\pm$ 3.8)} & 15.5 \footnotesize{($\pm$ 2.3)} & 28.0 \footnotesize{($\pm$ 1.8)} 
   & 5.1 \footnotesize{($\pm$ 2.4)} & 17.8 \footnotesize{($\pm$ 2.4)} & 26.4 \footnotesize{($\pm$ 1.2)} \\

    & \textsc{PTR} 
    & 14.7 \footnotesize{($\pm$ 1.1)} & 53.9 \footnotesize{($\pm$ 1.9)} & 80.6 \footnotesize{($\pm$ 1.2)}  
    & 8.6 \footnotesize{($\pm$ 2.5)} & 24.9 \footnotesize{($\pm$ 3.1)} & 30.7 \footnotesize{($\pm$ 2.0)}
    & 9.4 \footnotesize{($\pm$ 0.7)} & 26.9 \footnotesize{($\pm$ 1.5)} & 31.4 \footnotesize{($\pm$ 0.3)}  \\
    
    & {\ours} 
    & 28.6 \footnotesize{($\pm$ 6.2)} & 66.1 \footnotesize{($\pm$ 8.6)} & 80.9 \footnotesize{($\pm$ 1.6)}
    & 17.6 \footnotesize{($\pm$ 1.8)} & 28.8 \footnotesize{($\pm$ 2.0)} & 34.7 \footnotesize{($\pm$ 1.8)}
    & 17.8 \footnotesize{($\pm$ 2.2)} & 30.4 \footnotesize{($\pm$ 0.5)} & 33.2 \footnotesize{($\pm$ 1.4)}   \\
\cmidrule{2-11}

    & \textbf{\ourss}  
     & \textbf{33.3} \footnotesize{($\pm$ 1.6)}
     & \textbf{69.7} \footnotesize{($\pm$ 1.7)}
    & \textbf{81.8} \footnotesize{($\pm$ 1.0)}
    
    & \textbf{19.5} \footnotesize{($\pm$ 1.5)}
     & \textbf{30.7} \footnotesize{($\pm$ 1.7)}
    & \textbf{36.1} \footnotesize{($\pm$ 1.2)}
    
    & \textbf{18.7} \footnotesize{($\pm$ 1.8)}
     & \textbf{30.6} \footnotesize{($\pm$ 0.2)}
    & \textbf{35.3} \footnotesize{($\pm$ 0.3)}\\

\bottomrule
  
\end{tabular}
}

\end{table*}

\begin{table}[!htbp]
\small
\center
\caption{
Detailed Statistics for RE datasets.}
\label{tab:data_analysis} 
\scalebox{0.9}{
\begin{tabular}{l|r|r|r|c}
\toprule
Dataset & \textbf{\#~Train.} & \textbf{\#~Val.} & \textbf{\#~Test.} & \textbf{\#~Rel.} \\
\midrule
SemEval &6,507 &1,493 &2,717 &19\\
DialogRE & 5,963& 1,928& 1,858&36\\
TACRED &68,124 &22,631 &15,509 & 42\\
TACREV &68,124 &22,631 &15,509 & 42\\
Re-TACRED &58,465 &19,584 &13,418 &40\\
\bottomrule
\end{tabular}
}
\end{table}

\section{Experiments} 

\subsection{Dataset and Baselines}
We evaluate our model on five RE datasets:
SemEval 2010 Task 8 (SemEval)~\cite{hendrickx2010semeval},
DialogRE~\cite{DBLP:conf/acl/YuSCY20},
TACRED~\cite{zhang2017position},
TACREV~\cite{alt2020tacred}, Re-TACRED~\cite{stoica2021re}. 
Statistical details are shown in Table~\ref{tab:data_analysis}.

\subsection{Implementation Details}
We adopt Roberta-large as weight initialization (except to DialogRE with Roberta-base for a fair comparison).
We set k to be 16, $\lambda$ to be 0.2. As for the training parameters, we assign max sentence length as 128, Adam learning rate as 3e-5, and dropout rate as 0.1.

\textbf{Standard Setting.}
We leverage full trainsets to conduct experiments and compare with several knowledge-enhanced models, including
\textsc{SpanBERT}~\cite{joshi2020spanbert}, \textsc{KnowBERT}~\cite{peters2019knowledge}, \textsc{LUKE}~\cite{yamada2020luke}, and \textsc{MTB}~\cite{baldini-soares-etal-2019-matching}, which are  strong baselines for our method. 
We also compare with previous prompt tuning methods  
in the standard supervised setting.

\textbf{Low-Resource Setting.}
Unlike the 8-, 16-, and 32-shot settings in PTR \cite{ptr} and KnowPrompt \cite{chenKnowprompt2022}, we conduct experiments with settings of 1-, 5-, and 16-shot experiments to evaluate whether our method can achieve superior performance in extreme low-resource scenarios.
We randomly sample data five times using a fixed set of seeds, then record the average performance and variance.
For the process of sampling, we sample $k$ instances of each relation labels from the initial training  sets to form the few-shot training sets.

\begin{table}[!htb]
\center
\small
\caption{Standard RE performance of $F_1$ scores (\%) on different test sets. 
Best results are bold.}
\label{tab:supervised}
\scalebox{0.8}{
\begin{tabular}{l|c|c|c|c|c}
\hline
\toprule
\multicolumn{6}{c}{\textit{Standard Supervised Setting}}\\
\midrule
Methods    & SemEval  & DialogRE & TACRED 
& TACREV & Re-TACRED  \\
\midrule
\multicolumn{6}{c}{Fine-tuning pre-trained models}               \\
\midrule
\textsc{Fine-tuning}~  & 87.6 &57.3 & 68.7 & 76.0 & 84.9     \\

\textsc{SpanBERT}~\cite{joshi2020spanbert}      
 & - & - 
& 70.8 & 78.0  & 85.3        \\
\textsc{KnowBERT}~\cite{peters2019knowledge}
  & 89.1  & -  
& 71.5 & 79.3  & 89.1     \\
\textsc{LUKE}~\cite{yamada2020luke}   & -   & - 
& \textbf{72.7} & 80.6  & -         \\
\textsc{MTB}~\cite{baldini-soares-etal-2019-matching}   & 89.5   & -  & 70.1 & - & -      \\
\textsc{GDPNet} ~\cite{Fuzhao2021gdpnet}
 & -  & 64.9 
& 71.5 & 79.3  & -       \\
\textsc{Dual} ~\cite{semdialogre}
   & -  & 67.3 & - & -  & -      \\

\midrule
\multicolumn{6}{c}{Prompt-tuning pre-trained models}               \\


\midrule
\textsc{PTR}~\cite{ptr}   & 89.9  & 63.2 & 72.4 & 81.4   & 90.9     \\

\textsc{\ours}
& {90.2} 
& {68.6}  
& 72.4  
& {82.4}  
& {91.3}  
    \\

\textbf{\textsc{\ourss}}
& \textbf{90.4} 
& \textbf{69.4}  
& \textbf{72.7} 
& \textbf{82.7}  
& \textbf{91.5}  
    \\
    
\bottomrule
\hline
\end{tabular}
}
\end{table}

\begin{figure}
\hspace{-10pt}
    \centering
    \subfigure[{$\lambda$} varies.]{
    \includegraphics[width=0.22\textwidth]{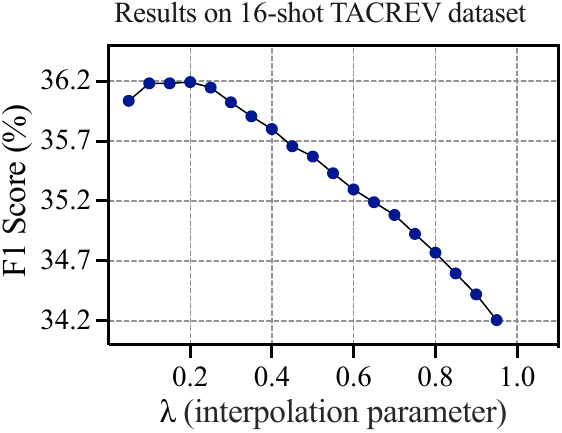}
    \label{fig:analysis1}
    }
    \hspace{-15pt}
    \quad
    \subfigure[{$k$} varies.]{
     \includegraphics[width=0.22\textwidth]{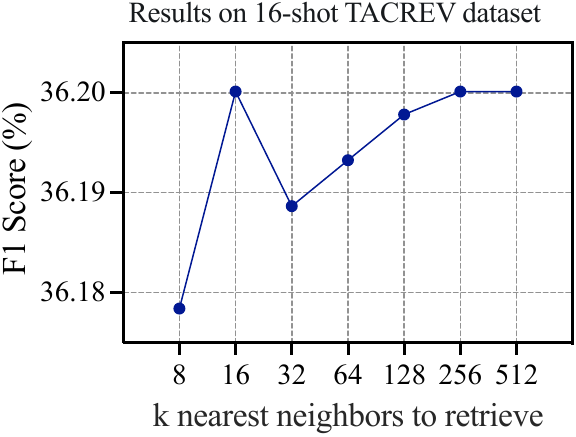}
     \label{fig:analysis2}
     }
     \caption{\label{fig:analysis} Effect of the parameters of the retrieval process.
    }
\end{figure}

\subsection{Main Results}
As shown in Table~\ref{tab:low-resource} and Table~\ref{tab:supervised},
we can observe that our {\ourss} achieves improvements over all baselines and even perform better than knowledge-enhanced models and prompt tuning methods in the standard supervised setting.
Table~\ref{tab:low-resource} shows that our {\ourss} can also achieve improvements of F1 scores when the training data is exceptionally scarce.
Note that our {\ourss} not only performs better in F1 scores but also achieves lower variance than PTR and KnowPrompt.
This finding proves that open-book examination with a retrieval component effectively enhances the prompt tuning since the knowledge learned in the weights of PLMs may be insufficient and unstable in an extreme low-resource setting.

\subsection{Analysis}

\subsubsection{Effect of  Interpolation Ratio}
Since the final distribution is incorporated by the PLM's and retrieved distribution, 
we apply analysis to view the impact of interpolation ratio parameter $\lambda$ for the final results.
From Figure \ref{fig:analysis1}, we find that model achieves optimal results on a 16-shot TACREV dataset when $\lambda$ is set to be  0.2. 
The suitable $\lambda$ can help the model correct the wrong prediction.

\begin{table}[!htbp]
    \begin{center}
    \small
    \caption{Ablation study  on TACRED.}
    \label{tab:ablation}
    \scalebox{0.85}{
    \centering
    \begin{tabular}{l c c c c}
    \toprule
    \tf{Method}
    & \tf{K=1} & \tf{K=5} & \tf{K=16} & \tf{Full} \\
    \toprule
    \textsc{\ourss}    & \tf{19.5}   &\tf{30.7}  & \tf{36.1}      &\tf{72.7}   \\
    \midrule
    None Retrieval  
    & 17.2  & 28.3
    & 34.1  & 72.2   \\
    Retrieval (\tt{[CLS]})
    & 17.4  & 28.6
    & 34.3  & 72.0   \\
    Retrieval (Sentence-BERT) 
    & 18.0  & 29.1
    & 34.9  & 72.4   \\
    
    Retrieval (TF-IDF) 
    & 16.8  & 27.7
    & 34.2  & 72.1   \\
 
    \bottomrule
    \end{tabular}
    }
    \end{center}
\end{table}

\subsubsection{Effect of $k$ Number for Neighbors}
We conduct experiments to validate the impact of the different number of $k$ nearest neighbor for performance.
From Figure~\ref{fig:analysis2}, we notice that the model performance continues to improve as $k$ increases until it converges when reaching a threshold ($k=16$).
We think this is because too many instances for retrieval may have a confusing influence on the prediction.

\subsubsection{Analysis of Efficiency}
To search over the open-book datastore, we adopt a fast nearest neighbor retrieval library FAISS~\cite{DBLP:journals/tbd/JohnsonDJ21}.
Note that the max number of samples on the training set is less than 70,000 in the standard supervised setting; thus, the increased time complexity is acceptable if $k$ holds. 
We make the comparison between KnowPrompt and {\ourss} in speed on TACRED for $k=16$.
We find that {\ourss} performs approximately 1.34 and 1.12 times slower than KnowPrompt in the standard supervised and few-shot setting, respectively.

\subsection{Ablation Study for  Retrieval Component}
We introduce the variants  of our model as follows:
1) \textit{None Retrieval}: we ablate the retriever module from our method. Thus, the model is downgraded to a regular prompt tuning model for RE; 
2) \textit{Retrieval (\tt{[CLS]})}: we construct the datastore with the hidden vectors from the position of  \texttt{[CLS]} and adopt the representation of \texttt{[CLS]} of instance at test time;
3) \textit{Retrieval (Sentence-BERT)}: we leverage additional Sentence-BERT~\cite{sentencebert}  to construct datastore and conduct query at  test time;
and 4) \textit{Retrieval (TF-IDF)}: we utilize TF-IDF to find and rank the nearest neighbour of instance in at  test time.
From Table~\ref{tab:ablation}, we find that the performance of the four variants drops a lot, which proves the effectiveness of our retrieval component and the advantage of our prompt-based instance representation.

\subsection{Case Study}

\begin{figure}[!htbp] 
\centering 
\includegraphics[width=0.45\textwidth]{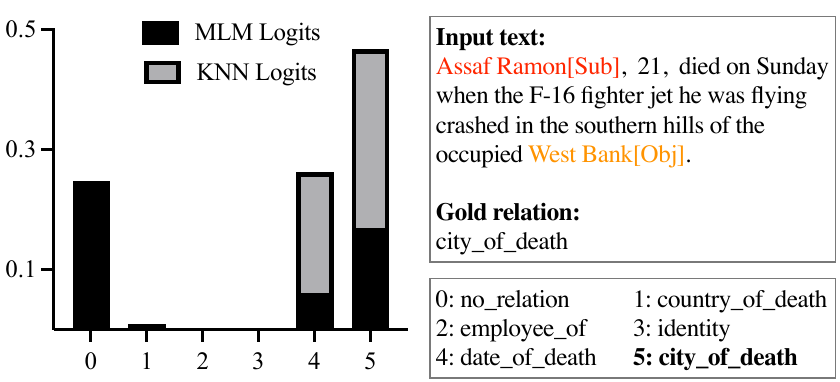} 
\caption{Case study. 
We omit most of the relation on the horizontal axis in the histogram due to space limitations.} 
\label{fig:case}
\end{figure}

We argue that the existing memorization-based inference paradigm for  RE cannot perform well for those scarce relations with minor instances.
Thus, we conduct case studies to analyze how our method helps the model reference the training set for calibrating the predicted distribution of the model.
Specifically, we choose the example with scare class from TACREV and visualize its calibration process. 
Note that we omit most of the relation types with low probability in the histogram of Figure~\ref{fig:case} due to space limitations.
We observe that MLM logits have high value on the relation ``\textit{no\_relation}''  while retrieval-based logits have high value on the relation ``\textit{city\_of\_death}''.
Our retrieval component effectively enhances the prompt tuning since the knowledge learned in the weights of PLMs may be insufficient and unstable.

\section{Related Work}
\subsection{Relation Extraction}

Early approaches of relation extraction involve pattern-based methods~\citep{DBLP:conf/ijcai/Huffman95}, CNN/RNN-based~\citep{DBLP:conf/acl/ZhouSTQLHX16} and graph-based methods ~\cite{guo2020learning}.
More recent works utilize  PLMs~\cite{DBLP:journals/corr/abs-1907-11692,devlin2018bert} for RE tasks~\cite{DBLP:conf/naacl/ZhangDSWCZC19,www20-Zhou,wang2020finding,generative_triple,docunet,li2020logic,AliCG,PRGC}; however, such a paradigm is still suboptimal due to the gap between pre-training and downstream tasks.
Thus, a series of prompt tuning methods for RE have been further explored, showing significant improvement in widespread benchmarks.
Typically, PTR~\cite{ptr}  applies logic rules in prompt-tuning  with
several sub-prompts and encode prior knowledge. KnowPrompt~\cite{chenKnowprompt2022}  incorporates knowledge among relation labels into prompt tuning for RE and present synergistic optimization for better performance.

\subsection{Retrieval-based Language Models}
Recently, Retrieval-based language models (R-LMs) \cite{DBLP:journals/corr/abs-2201-12431} have risen to improve over language models in various tasks such as unconditional language modeling~\cite{DBLP:conf/iclr/KhandelwalLJZL20,he2021efficient}, machine translation~\cite{DBLP:conf/iclr/KhandelwalFJZL21,DBLP:conf/aaai/GuWCL18}, text classification~\cite{li2021knnbert} and question answering~\cite{DBLP:conf/emnlp/KassnerS20}. 
R-LMs is a non-parametric method that uses training examples to augment the language model predictions at test time.
The core ingredient is that they don't have to only rely on the knowledge encoded in the model's already-trained weights but use nearest neighbors to find augmented samples based on the untrained PLMs.
Different from above methods, we initially propose to build an open-book datastore based on already-trained PLMs with prompt tuning.
We are the first retrieval-enhanced prompt tuning approach for RE to the best of our knowledge.

\section{Conclusion}
In this work, we view RE as an {\it open-book examination} and propose retrieval-enhanced prompt tuning, a new paradigm for RE that allows PLMs to reference similar instances from open-book datastore.
The success of \ourss~suggests that retrieving the related contexts based on prompt tuning as references makes it more accessible for the PLMs to predict long-tailed or hard patterns.
We will consider enriching the composition of the open-book datastore with more specific features in future work.

\begin{acks}
We  want to express gratitude to the anonymous reviewers for their hard work and kind comments. 
This work is funded by NSFC 91846204/U19B2027, National Key R\&D Program of China (Funding No.SQ2018YFC000004), Zhejiang Provincial Natural Science Foundation of China (No. LGG22F030011), Ningbo Natural Science Foundation (2021J190), and Yongjiang Talent Introduction Programme (2021A-156-G). Our work is supported by Information Technology Center and State Key Lab of CAD\&CG, ZheJiang University.
\end{acks}

\bibliographystyle{ACM-Reference-Format}
\balance
\bibliography{sample-base}


\end{document}